\documentclass[10pt,twocolumn,letterpaper]{article}

\usepackage{cvpr}
\usepackage{times}
\usepackage{epsfig}
\usepackage{graphicx}
\usepackage{amsmath}
\usepackage{amssymb}
\usepackage{booktabs}

\usepackage[normalem]{ulem}
\usepackage[utf8]{inputenc} 
\usepackage[T1]{fontenc}    
\usepackage{url}            
\usepackage{booktabs}       
\usepackage{amsfonts}       
\usepackage{nicefrac}       
\usepackage{microtype}      
\usepackage{amsmath}
\usepackage{amsthm}
\usepackage{amssymb}

\usepackage{algorithm}
\usepackage[noend]{algpseudocode}
\usepackage{graphicx}
\usepackage{eqparbox}
\usepackage{placeins}
\usepackage{mathtools}
\usepackage{color,enumitem}

\usepackage[pagebackref=true,breaklinks=true,letterpaper=true,colorlinks,bookmarks=false]{hyperref}

\cvprfinalcopy 

\def\cvprPaperID{473} 

\ifcvprfinal\fi
\begin{document}

\title{A Two-Step Disentanglement Method}

\author{Naama Hadad\\
Tel Aviv University\\
\and
Lior Wolf\\
Facebook AI Research and Tel Aviv University\\
\and
Moni Shahar\\
Tel Aviv University
}

\maketitle

\begin{abstract}
We address the problem of disentanglement of factors that generate a given data into those that are correlated with the labeling and those that are not. Our solution is simpler than previous solutions and employs adversarial training. First, the part of the data that is correlated with the labels is extracted by training a classifier. Then, the other part is extracted such that it enables the reconstruction of the original data but does not contain label information. The utility of the new method is demonstrated on visual datasets as well as on financial data. Our code is available at~\url{https://github.com/naamahadad/A-Two-Step-Disentanglement-Method}.
\end{abstract}

\section{Introduction}


The problem of identifying complementary factors and separating them from each other is ubiquitous. In face recognition and in object recognition, one would like to separate illumination and pose from identity or label. In handwriting recognition, we would like to separate the factors that define the content of the text written from those that define its style. The separation between what is spoken and who is the speaker in automatic speech recognition and multi-speaker speech synthesis is similar in nature. In each of these domains, specialized solutions have emerged, oftentimes emphasizing recognition and eliminating the other factors and sometimes employing compound labels from orthogonal domains. However, the task of separating the factors that generated the observations, which is called \textit{disentanglement}, is also being studied as an abstract pattern recognition problem.

In this work, we present a new algorithm for disentanglement of factors,  where the separation is based on whether the factors are relevant to a given classification problem. Following the terminology used in~\cite{disentanglement}, we call the factors that are relevant for the classification task \textit{specified factors}, and those which are not \textit{unspecified factors}. 
In order to perform disentanglement, we present a new adversarial technique. First, a classifier $S$ is trained to predict the specified factors. The activations of $S$ are then used to capture the specified component of the samples. A second network $Z$ is then trained to recover the complimentary component. A first loss on $Z$ ensures that the original sample can be reconstructed from the output of both networks together ($S$ and $Z$). A second loss on $Z$, which is based on an adversarial network, ensures that $Z$ does not encode the specified factors. The algorithm has the advantage that is makes very weak assumptions about the distribution of the specified and the unspecified factors. 

We focus our experiments on the image-based benchmarks used in previous work. In addition to image data, we also evaluate our model on financial data. A simplified model for stock price changes is that the price change can be decomposed into two factors, market movement and idiosyncratic movement. A common assumption is that the market return is a Geometric Brownian motion and cannot be predicted. However, since different stocks have different correlations with the market, one can neutralize the market factor from her portfolio. In real-life trading, the correlations are non-stationary and there are other effects such as trading costs that should be taken into account. Despite all that, the disentanglement of driving factors is relevant both for prediction purposes as well as for data generation. 

\subsection{Related work}

\noindent {\bf Disentanglement} was studied in many contexts and has a vast literature. Early attempts to separate text from graphics using basic computer vision tools were made in~\cite{fletch}. In~\cite{tenenbaum} voice data was analyzed. It was assumed that the data was generated by two sources and separation was done using a bilinear model. Manifold learning methods was used by ElGammal and Lee in order to separate the body configuration from the appearance~\cite{elgammal}. In recent years, few papers tackled this problem using neural networks. What-where encoders~\cite{huang} combine the reconstruction criteria with the discrimination in order to separate the factors that are relevant for the labels. In~\cite{kingma} variational auto encoders were used to separate the digit from the style. However their approach can not generalized to unseen identities. This restriction was relaxed in~\cite{disentanglement}, where they trained a conditional generative model by using an adversarial network to remove label information from the unspecified part of the encoding.  

Concurrently with our work, the Fader Networks~\cite{fader} employ an architecture that is closely related to the second step of our two-step architecture. While in our model a classifier is trained to capture the specified factors, in the architecture of~\cite{fader}, the labels are used directly. The main advantage of our architecture in comparison to the one step alternative, is its support of novel labels at test time, i.e., it is not limited to the set of labels seen during training. This quality of our architecture is crucial for the Norb and Sprites datasets we present later, where we use the disentanglement for new identities at test time. In the modeling of the financial data this quality also comes into effect. For this data, the specified factors (the labels) denote the market regime during train years, whereas during test years there may be different market regimes.

\noindent{\bf Generative Adversarial Networks} GAN~\cite{gan} is a method to train a generator network $G$ that synthesizes samples from a target distribution given noisy inputs. In this approach, a second network called the discriminator $D$ is jointly trained to distinguish between generated samples and data samples. This ``competition'' between the generator and the discriminator, induces a zero-sum game whose equilibrium is reached when the discriminator can no longer distinguish between the generated samples and the empirical ones. Since this approach was published, many variations on this idea has appeared, see for example~\cite{radford,denton2015deep,infogan}. 

\section{Method}

\paragraph{The Problem of Disentanglement}
We are given a set of labeled inputs $X$ with the matching labels $Y$. Our goal is to represent the data using two disjoint parts of a code, $S$, and $Z$. We require $S$ to contain all the information relevant for the class ids $Y$, and $Z$ to contain only the unspecified factors of the data. For the example of handwriting recognition, if $Y$ is the text written in the image samples $X$, then $S$ will contain the information about the textual content $Y$, whereas $Z$ will only contain information about the style of writing. 

\begin{figure}[t]
  \centering
      \begin{tabular}{c}
  \includegraphics[width=.7\linewidth,trim={0 1cm 0 0},clip]{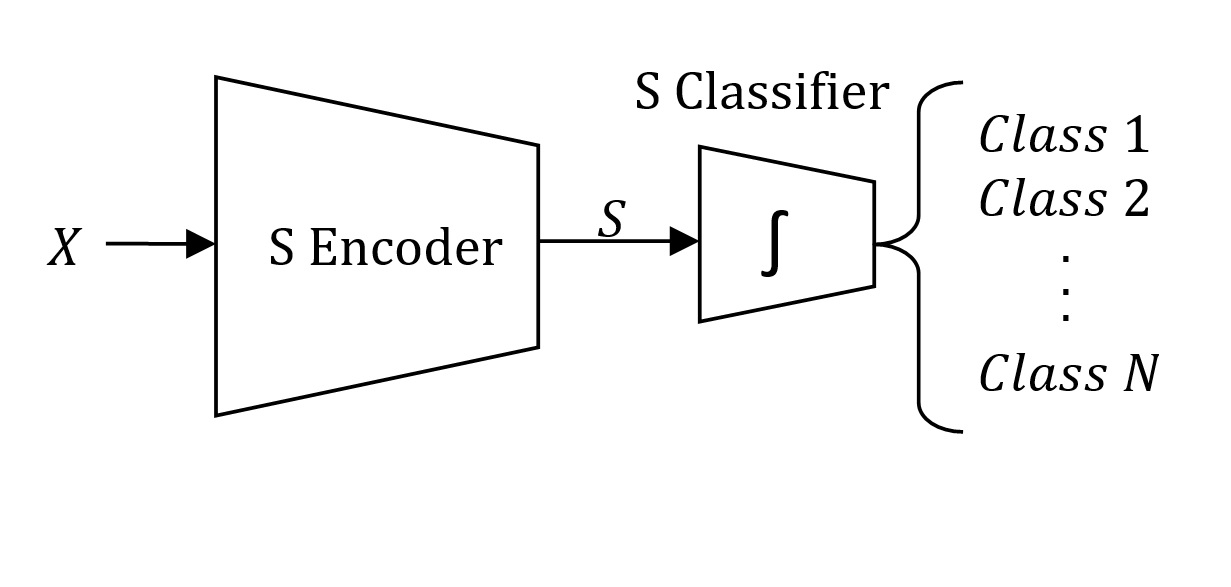} \\
    (a) \\
\includegraphics[width=.9\linewidth,trim={0 1cm 0 0},clip]{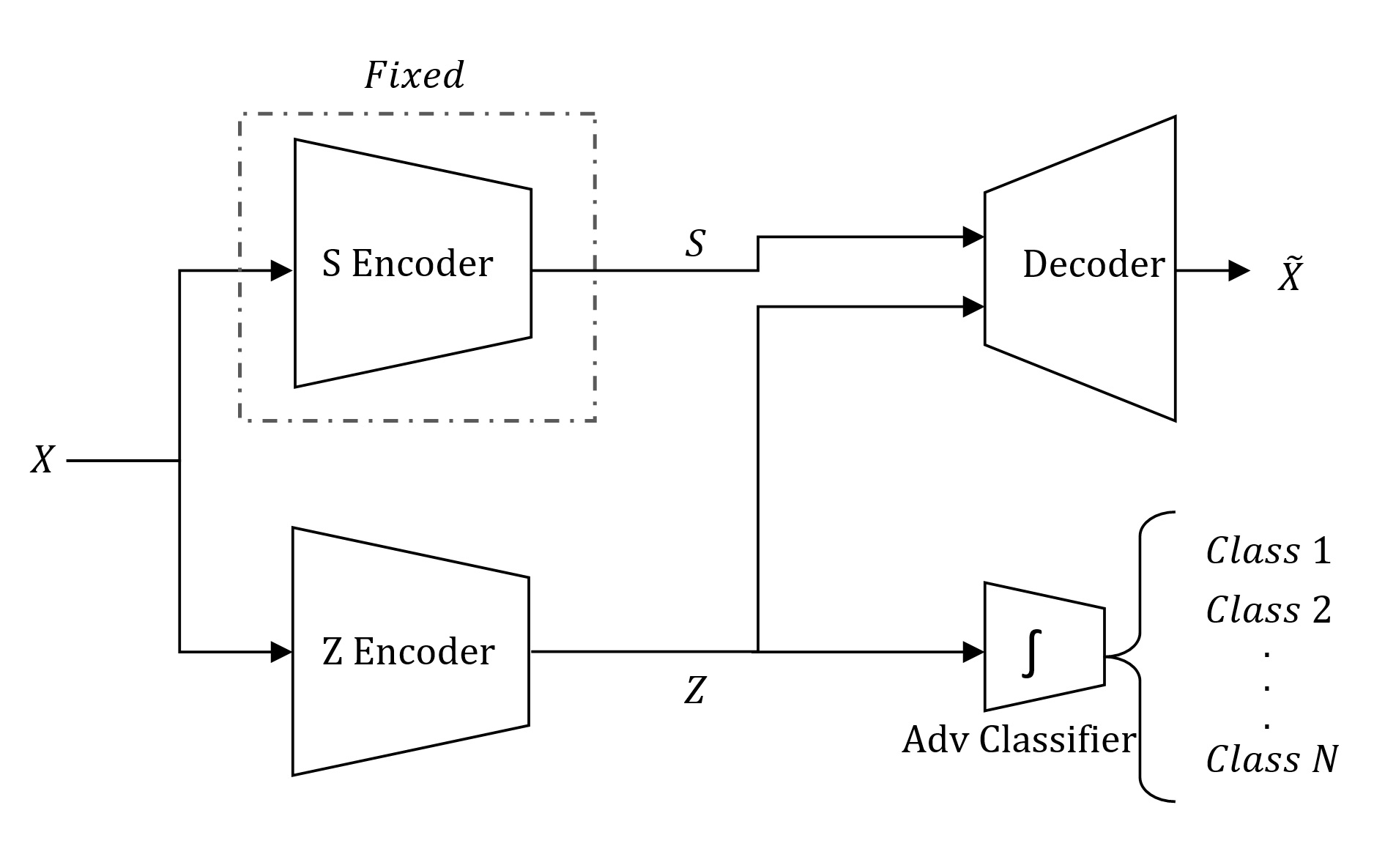} \\
    (b)\\
    \end{tabular}
  \caption{\label{fig0_Net} Network architecture: (a) We train the  $S$ encoder and its classification network on a pure classification task. (b) Once $S$ is given, we freeze its weights and train the Enc-Dec network and the adversarial classifier alternatively}
\end{figure}

\paragraph{The Model}
For the encoding, we chose $S$ and $Z$ to be vectors of real numbers rather than a one-hot vector. This idea, presented in~\cite{disentanglement}, enables the network to generalize to identities that did not appear in the training set.  

We define a new network architecture for the disentanglement of the factors. It is simpler and more straightforward than the one presented in~\cite{disentanglement}. The network contains two deterministic encoders to map $X$ to its specified and unspecified components $S=Enc_S(X)$ and $Z=Enc_Z(X)$ accordingly. To train the $S$ encoder $Enc_S$, we first use a sub-network for the classification task and train the $S$-classifier concurrently with $Enc_S$. This sub-network accepts $X$ as its input, encodes it to a vector $S$, and then runs the $S$-classifier on $S$ to obtain the labels $Y$, see Fig.~\ref{fig0_Net}(a). The result of this network is an encoding of the data that contains the information needed in order to predict the class identity. 

In a second step $Enc_S$ is kept fixed. To train the $Z$-encoder to ignore the specified factors and contain data only on the unspecified factors, we use a new variation of adversarial networks. The configuration of the network is given in Fig~\ref{fig0_Net}(b), and it is composed out of two network branches. The adversarial classifier (see the bottom part of the figure) is being trained to minimize the classification loss given $Z,Y$ as input, namely, it is trained to classify $Z$ to $Y$. The Enc-Dec network (the rest of the network) is trained to minimize the sum of two terms: (i) the reconstruction error (given $S$ and $Z$), and (ii) minus the adversarial network loss.

More formally, let $\theta_Z$ be the parameters of $Enc_Z(X)$ and let $\theta_X$ be the parameters of the reconstruction network with output $\tilde{X} = Dec(S,Z)$. Let $\theta_A$ be the parameters of the adversarial network. We define $L_{adv}(\{(Z,Y)\},\theta_A)$ to be the classification loss of the adversarial network and $L_{rec}(\{S,Z,X\},\theta_X)$ to be the reconstruction loss of $\tilde{X}$. When optimizing $\theta_A$, $L_{adv}$ is minimized. When optimizing the two other networks, $\theta_Z$ and $\theta_X$, the objective is to simultaneously minimize $L_{rec}$ and maximize $L_{adv}$. Hence, our objective is: 

\begin{equation}\label{eqn_loss}
	\underset{\theta_Z,\theta_X,\theta_A}{min} \{L_{rec} - \lambda *L_{adv}\}, \lambda>0
\end{equation}

Note that while GANs are typically used in order to improve the quality of generated output such as images, here we use an adversarial configuration to encourage the encoding to "forget" information about the labels, which, in turn, leads to the disentanglement. 

Training the $S$ encoder together with the $Z$ encoder and the subsequent decoder, could lead the network to converge to a degenerated solution, where all information is encoded in $S$, whereas $Z$ holds no information on any factor. By training the $S$ network in the first stage with a limited capacity, and then fixing the values of its parameters, this scenario is avoided. Since $S$ has a limited capacity it ignores most of the information on the unspecified factors, which is irrelevant for its goal.  
\paragraph{Training details}

We employ MSE for the $L_{rec}$ loss, and use categorical cross-entropy loss for both the $S$ classifier's loss and $L_{adv}$. The $\lambda$ for each dataset was chosen independently using few iterations on validation data. 

For the training of the $S$-network and the Enc-Dec network, we apply the Adam optimization method~\cite{adam} with a learning rate of 0.001 and beta of 0.9. For the adversarial classifier, we used SGD with a learning rate of 0.001. 

While training the $Z$-network, we have noticed that the adversarial part requires more steps to stabilize, since it should solve a complicated classification task on a changing input. 
Therefore, we run, at each iteration, one mini-batch to train the Enc-Dec network, followed by three mini-batches to train the adversarial network. 

\subsection{Comparison to~\cite{disentanglement} on Toy Data}
To illustrate the advantages of our approach in comparison to the more involved method of~\cite{disentanglement}, we generated images of a gray rectangle in ten possible locations on a white or black background. We refer to the background color as the unspecified factor $Z$, whereas the location of the rectangle is the specified factor $S$. We denote the ten possible values of $S$ by $\{s_0,\dots,s_9\}$. All possible twenty images were drawn with equal probability. 

 We also generated similar images, where the unspecified factor consists of two binary variables - the first controls the upper half background color and the second controls the lower half background color. Where similar to the first case all forty images were drawn with equal probability. 
 
We refer to the sets as \textit{Synth1} and  \textit{Synth2}. 
For the encoding, we chose both $S$ and $Z$ to be vectors of size $4$. We run our network and the network in~\cite{disentanglement} to obtain $S,Z$ for \textit{Synth1} and  \textit{Synth2}. We then used a neural network classifier with three dense layers of size $8$ on $S$ to find the label. The obtained accuracy was $100\%$ for both networks. 

\begin{figure}[t]
  \centering
    \begin{tabular}{c}
  \includegraphics[width=.948\linewidth]{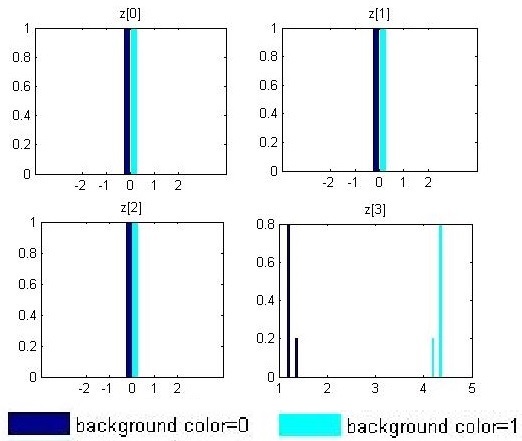}\\
  (a)\\
  \includegraphics[width=.948\linewidth,trim={0 1cm 0 0},clip]{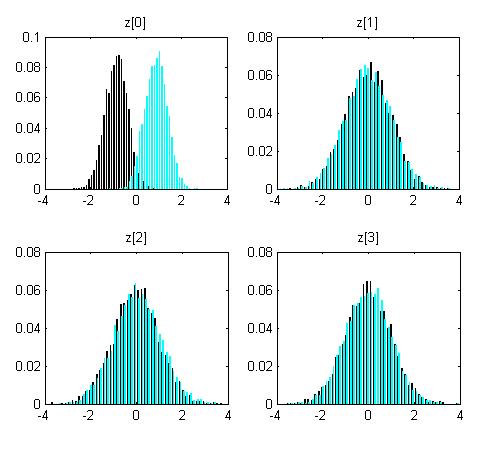}\\
    (b)\\
    \end{tabular}
  \caption{\label{fig_Synt1} {\it Synth1} data: The dimension of the unspecified factors is $1$. (a) histogram of different $Z$ components of our model (b) same histogram for the encoding of~\cite{disentanglement}}
\end{figure}

\begin{figure}[t]
  \centering
    \begin{tabular}{c}
  \includegraphics[width=.948\linewidth]{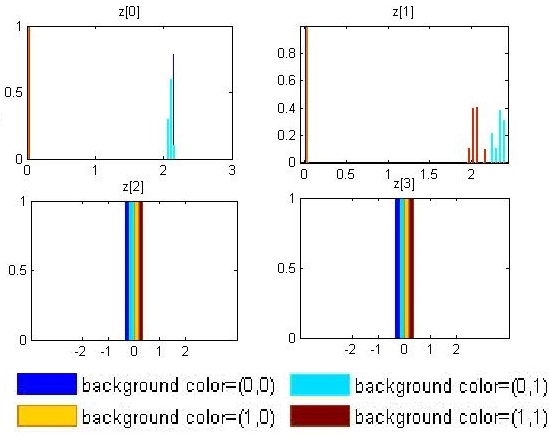}\\
  (a)\\
  \includegraphics[width=.948\linewidth]{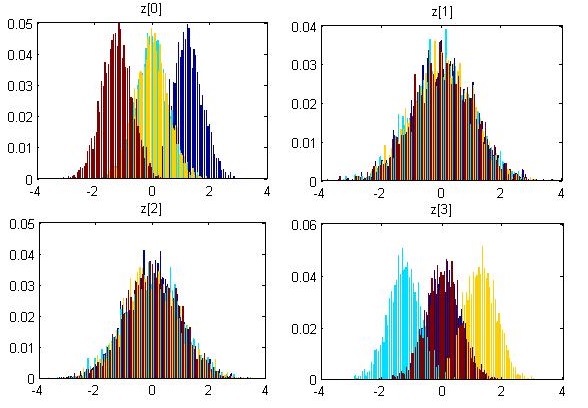}\\
    (b)\\
    \end{tabular}
  \caption{\label{fig_Synt2} {\it Synth2} data: The dimesnion of the unspecified factors is $2$. (a) An histogram of the components of $Z$ of our model. (b) Same histogram for the encoding from the model of~\cite{disentanglement}}
\end{figure}

We then examined the $Z$ vectors that were obtained from both methods of disentanglement on \textit{Synth1} and \textit{Synth2}. First we verified using a classifier that the background color can be inferred from $Z$ perfectly, whereas inferring the location of the rectangle from $Z$ leads to accuracy that is close to random. Next we turned to examining the distribution of $Z$ for \textit{Synth1} and \textit{Synth2}, these distributions are presented in Fig.~\ref{fig_Synt1},\ref{fig_Synt2} respectively. The figures show the values of the components $Z_0,\dots,Z_3$ for all of the data points. Each color represents a different value of the latent variable $Z$. Note that since the method in~\cite{disentanglement} defines $Z$ to be a random vector drawn from a normal distribution specified by $\mu,\sigma$, we show a sample of the drawn $Z$ vectors. 

For the binary case (\textit{Synth1}, Fig.~\ref{fig_Synt1}) our encoding shows two narrow peaks well separated in $Z_3$, whereas all other components have one peak 
(the coordinate that contains the information is completely arbitrary since all coordinates are treated the same way). In the VAE encoder, the information was separated only for $Z_0$, but even the best classifier (which is LDA in this case) will have some error, since the peaks were not disjoint. This simple experiment also demonstrates that our results are simpler to understand. 

The gap in the explicitness of the results as encoded in $Z$  is more apparent on \textit{Synth2}. In Fig.~\ref{fig_Synt2}, we see that our encoding of $Z$ is well separated on $Z_0$ and $Z_1$ while in the other method, one cannot tell the number of values without further analysis or prior knowledge about $Z$. Moreover, applying standard PCA on the sampled $Z$ vector of the auto encoder, gave four components with similar variance, as shown in Tab.~\ref{tbl_syntpca}.

\begin{table}[]
\centering
\begin{tabular}{ccccc}
\toprule
       & \multicolumn{4}{c}{\textbf{\textit{Synth1}}}                       \\ \cmidrule{2-5} 
\textbf{PCA component:}             & \textbf{1} & \textbf{2} & \textbf{3} & \textbf{4} \\ \midrule
Our model             & 1.000      &            &            &            \\
~\cite{disentanglement}  & 0.252      & 0.251      & 0.25       & 0.247 \\ 
\cmidrule{2-5}       & \multicolumn{4}{c}{\textbf{\textit{Synth2}}}                       \\ 
\textbf{PCA component:}             & \textbf{1} & \textbf{2} & \textbf{3} & \textbf{4} \\ \midrule
Our model             & 0.610      & 0.390      &            &            \\
~\cite{disentanglement}  & 0.263      & 0.249      & 0.248      & 0.240      \\ 
\hline
\end{tabular}
\smallskip
\caption{The ratio of the variance of the $Z$-encoding projected on its PCA components. Note that the autoencoder split the information between all the components, whereas our encoding expressed the information using the minimum number of dimensions.}
\label{tbl_syntpca}
\end{table}

\section{Experiments}

We evaluate our method on the visual disentanglement benchmarks used in previous work, as well as on simulated and real financial data. The detailed network architecture used for each of the experiments is described in Tab.~\ref{tab:tblarch}.

\begin{table}[t]
\centering
\resizebox{.995\linewidth}{!}{
\begin{tabular}{b{1.6cm}lll}
\hline
& Image Datasets & \textbf{Stocks return} \\ \hline
\textbf{Encoders S,Z}                                                      & \begin{tabular}[c]{@{}l@{}}For MNIST and Sprites\\three 
5x5 convolotional, for \\ NORB and Extended YaleB three 3x3 \\ convolutional layers.\\ 
All convolutional layers with stride 2 \\ and a dense S/Z dimension layer. \\ all with ReLU non-linearities\end{tabular}& \begin{tabular}[c]{@{}l@{}}4 dense layers of sizes \\ 100,66,66,50 with ReLU \\ non-linearities\end{tabular}                              \\ \hline

\textbf{S classifier}                                                      & \begin{tabular}[c]{@{}l@{}} For MNIST and Sprites dense layers\\x 256 hidden units, \\ for NORB,Extended YaleB\\x 16 hidden units \\ Batch Normalization, ReLU\\ and  a softmax for the output\end{tabular}                                    & \begin{tabular}[c]{@{}l@{}}2 dense layers x\\ 50 hidden units, \\ Batch Normalization,\\ ReLU and a softmax output\end{tabular}  \\ \hline

\textbf{Decoder}                                                           & \begin{tabular}[c]{@{}l@{}}Mirroring network to the encoders: \\ dense layer and three convolutional\\ network with upsampling\end{tabular}                 & \begin{tabular}[c]{@{}l@{}}4 dense layers of sizes \\ 70,66,66,100 with\\ ReLU non-linearities\end{tabular}                              \\ \hline

\textbf{\begin{tabular}[c]{@{}l@{}}Adversarial \\ Classifier\end{tabular}} & \begin{tabular}[c]{@{}l@{}}3 dense layers x 256 hidden units, \\ Batch Normalization,ReLU \\ and a softmax for the output\end{tabular}                & \begin{tabular}[c]{@{}l@{}}3 dense layers x 50 \\ hidden units, \\ Batch Normalization, ReLU \\ and a softmax for the output\end{tabular} \\ \hline

\textbf{S, Z \#dims} & 
\begin{tabular}[c]{@{}l@{}}Mnist: 16,16, Sprites: 32,128, \\ NORB and Extended YaleB: 32,256\end{tabular}& 20,50\\ \hline
\end{tabular}
}
\smallskip
\caption{Networks architectures}
\label{tab:tblarch}
\end{table}

\subsection{Image Benchmarks}

We followed a previous work~\cite{disentanglement} and tested our model on four visual datasets - MNIST~\cite{mnist}, NORB~\cite{norb}, Sprites dataset~\cite{sprites} and the Extended-YaleB dataset~\cite{yaleB}.

For measures of performance on the visual datasets, we also followed the ones suggested in~\cite{disentanglement}. Note that all these measures are subjective.

\begin{itemize}[leftmargin=*]
\item \textit{Swapping} - In swapping, we generate an image using $S$ from one image, $I_1$, and $Z$ from a different image, $I_2$. In a good disentanglement, the resultant image preserves the $S$-qualities of $I_1$ and the $Z$-qualities of $I_2$.

\item\textit{Interpolation} - Interpolation of two source images is a sequence of images generated by linearly interpolating the $S$ and $Z$ components  of the two different sources. The measure is again done by visually judging the resultant images. i.e., we expect to see "more" of the look of the second image, the bigger its weight gets. Interpolation is done in both the $S$ space and the $Z$ space.

\item\textit{Retrieval} - To assess the lack of correlation between the $S$ and the $Z$ components, we perform a query based on either the $S$ part or the $Z$ part, where in each case we retrieve its nearest neighbors in the corresponding space. 

\item\textit{Classification Score} - In addition to the qualitative measures above, we quantify the amount of information on the class that each part of the code ($S$ and $Z$) contains on the class. Since measuring this directly is a difficult task, we approximate it by running a classification algorithm. A good disentanglement is such that when running the classifier on the $S$ part it gives high accuracy, whereas when running it on the $Z$ part it gives nearly random results. 

\end{itemize}

\begin{figure}[t]
  \centering
    \begin{tabular}{c}
  \includegraphics[width=.9948\linewidth,trim={0 1cm 0 0},clip]{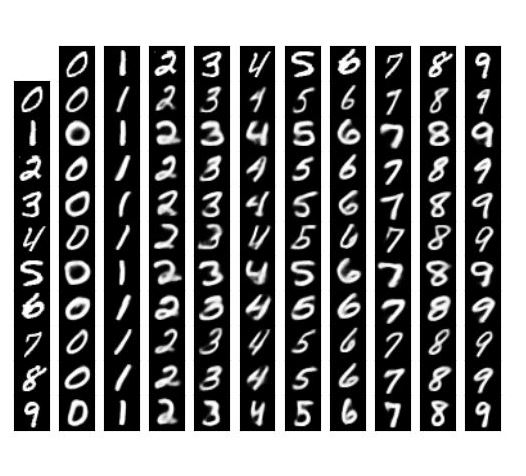}\\
  (a)\\
  \includegraphics[width=.9948\linewidth,trim={0 1cm 0 0},clip]{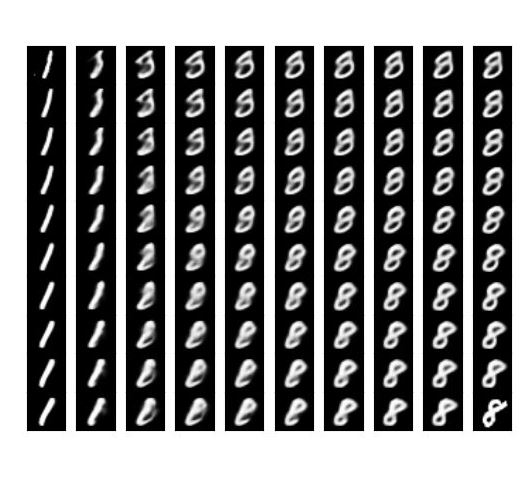}\\
    (b)\\
    \end{tabular}
  \caption{\label{fig1_Mnist}(a) Swapping the specified and unspecified components of MNIST images. The images are generated using Z from the left column and S from the top row in the decoder. The diagonal digits show reconstructions. (b) Interpolation results. the images in the top-left and bottom-right corners are from the test set. The other digits are generated by interpolation of S and Z gradually. Z interpolate along the rows and S through the columns. }
\end{figure}

\textbf{MNIST - } For the MNIST data, the $S$ part is the digit and the $Z$ part is the style. In Fig.~\ref{fig1_Mnist}, we present the results for swapping and interpolation. The rows of the table in the left hand side of the figure shows the style ($Z$) and the columns the digit. To the best of our judgment, the style looks well separated from the content.

\begin{figure}[t]
  \centering
  \begin{tabular}{c}
  \includegraphics[width=.948\linewidth,trim={0 1cm 0 0},clip]{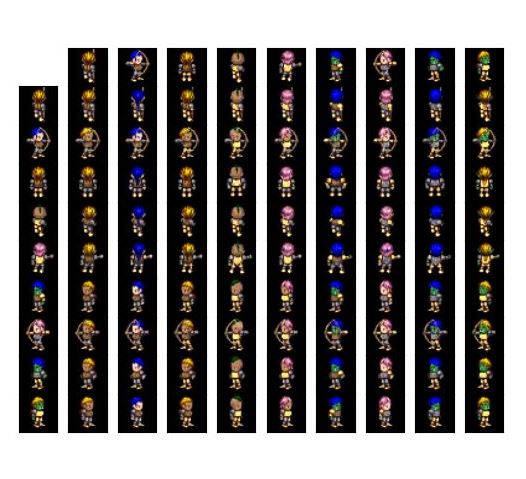}\\(a)\\
  \includegraphics[width=.948\linewidth,trim={0 1cm 0 0},clip]{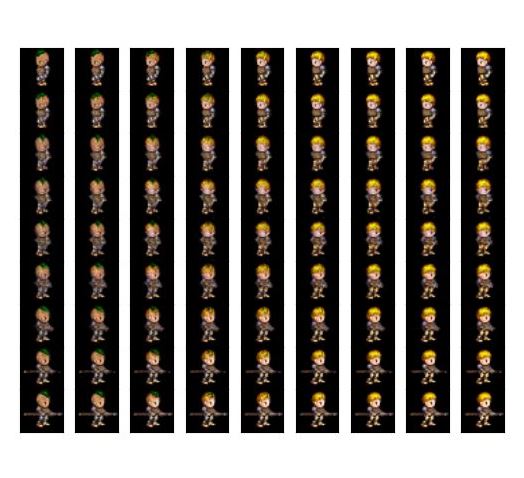}\\
    (b)\\
    \end{tabular}
  \caption{\label{fig2_Sprites} (a) Swapping the specified and unspecified components of Sprites. The images are generated using $Z$ from the left column and $S$ from the top row in the decoder. (b) Interpolation results. the images in the top-left and bottom-right corners are from the test set. The other images are generated by gradual interpolation of $S$ and $Z$. $Z$ interpolates along the rows and $S$ through the columns.}
\end{figure}

\textbf{Sprites dataset -} This dataset contains color images of sprites~\cite{sprites}\footnote{\url{http://lpc.opengameart.org/} (CC BY-SA 3.0).}. Each sprite character is defined by body type, gender, hair type, armor type, arm type and greaves type. Overall there are 672 different characters, from which we use 572 characters for the training set and 100 characters for the test set. For each character, there are five animations each from four viewpoints, each animation has between 6 and 13 frames. We use character's identity as the specified component. The results from swapping and interpolation are shown in Figure \ref{fig2_Sprites}. Our model learned to separate the character from its position and weapon and generalizes the separation to unseen characters.

\begin{figure}[t]
  \centering
  \begin{tabular}{c}
  \includegraphics[width=.948\linewidth,trim={0 1cm 0 0},clip]{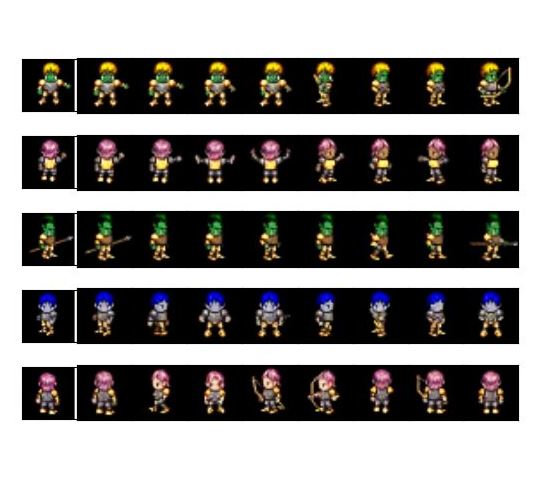}\\(a)\\
  \includegraphics[width=.948\linewidth,trim={0 1cm 0 0},clip]{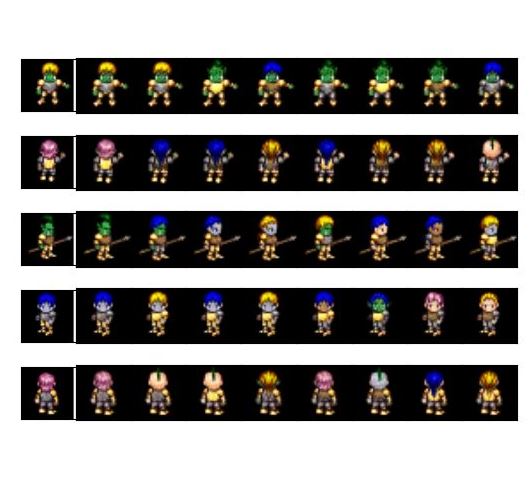}\\
    (b)\\
    \end{tabular}
  \caption{\label{fig3_Spritesret}Sprites retrieval results (a) Querying on the specified component S. (b) Querying on the unspecified component Z. The components of the sprites on the left column are used as the query. }
\end{figure}

Examining the retrieval results in Fig.~\ref{fig3_Spritesret}, it is possible to see that for the $Z$ part (sub-figure (b)), the characters in any row is random but its pose is kept. In the $S$ part (sub-figure (a)), the character is perfectly kept, whereas the pose is not. In~\cite{disentanglement}, it seems that $Z$ holds some information on $S$ because the hair style and color rarely changes between characters.

\textbf{Small NORB dataset~\cite{norb} -} The NORB dataset contains images of $50$ toys belonging to five generic categories: four-legged animals, human figures, airplanes, trucks, and cars. The objects were imaged by two cameras under six different illumination conditions, nine elevations and $18$ azimuths.
The training set is composed of five instances of each category and the test set of the remaining five instances. We use the instance identity as the specified component and have $25$ different labels for the training set. 

\begin{figure}[t]
  \centering
  \begin{tabular}{c}
    \includegraphics[width=.948\linewidth,trim={0 1cm 0 0},clip]{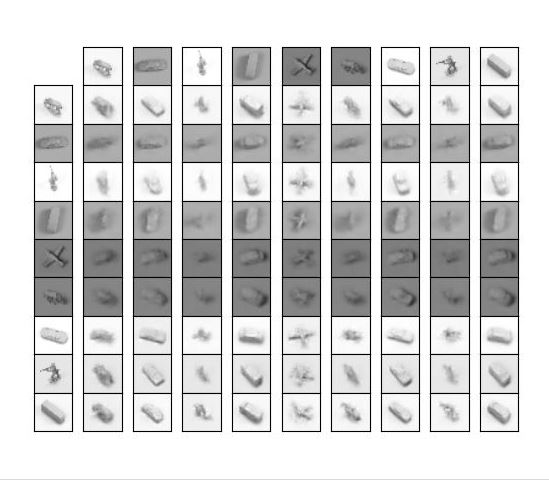}\\(a)\\
  \includegraphics[width=.948\linewidth,trim={0 1cm 0 0},clip]{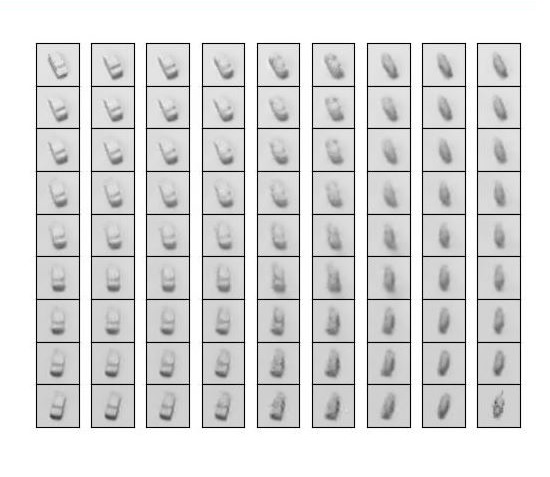}\\
    (b)\\
    \end{tabular}
  \caption{\label{fig3_norb} (a) Swapping the specified and unspecified components of the NORB test set images. (b) Interpolation results. These are the same arrangements as in Figure~\ref{fig2_Sprites}.}
\end{figure}

For this dataset the swapping results were not perfect. We succeeded in separating different azimuths and background from the instance.
However, for some of the categories, the reconstruction contained mistakes. This is probably due to the high variability between the instances in the train and the test. The numerical results support this hypothesis, since there are big difference between the train and the test errors. The results look good for the interpolation, see Figure~\ref{fig3_norb}. A similar degradation of results was also observed in~\cite{disentanglement}

\textbf{Extended-YaleB~\cite{yaleB} -} The extended-YaleB Face dataset contains $16095$ images of $28$ human subjects under nine poses and $64$ illumination conditions. 
The training (test) set contains $500$ (roughly $75$) images per subject. We use subject identity as the specified component. 

\begin{figure}[t]
  \centering
    \begin{tabular}{c}
    \includegraphics[width=.948\linewidth,trim={0 1cm 0 0},clip]{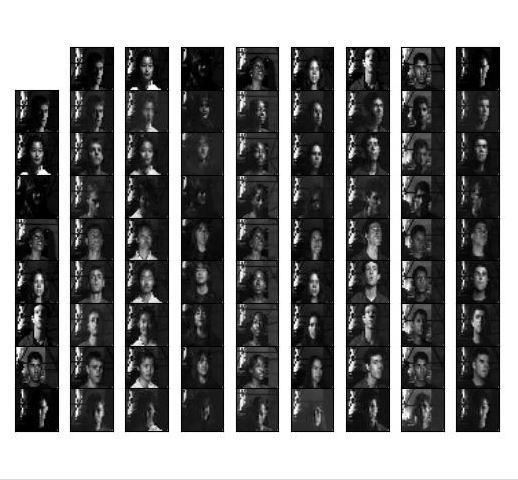}\\(a)\\
  \includegraphics[width=.948\linewidth,trim={0 1cm 0 0},clip]{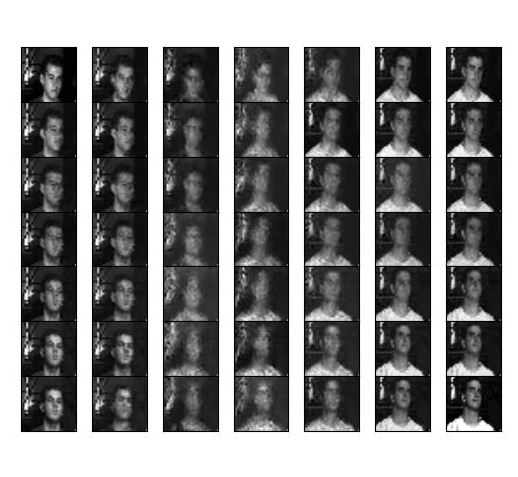}\\
    (b)\\
    \end{tabular}
  \caption{\label{fig5_yale} (a) Swapping the specified and unspecified components of Yale test set images. (b) Interpolation results. These are the same arrangements as in Figure~\ref{fig1_Mnist}. In this case, the results are visibly inferior to the examples presented in~\cite{disentanglement}}
\end{figure}

Results for swapping and interpolation for images from the test set shown in Figure~\ref{fig5_yale}. For swapping, one can see that illumination conditions are transferred almost perfectly, whereas the position is not perfectly transferred (see for example line 6, column 5). We again suspect that this is mainly because some of the positions were missing in the training set, and with more data we expect the results to improve. For the interpolation, some of the mixed identities do not resemble either sources. 

\paragraph{Quantitative results} 
The numerical results for all datasets are shown in Tab.~\ref{tbl1_numericresults}. One can see that the unspecified component is almost agnostic to the identity, while the classifier on the specified component achieves high accuracy.
For comparison with~\cite{disentanglement}, we added to the table the results that were reported in their paper.  For most cases our model achieves higher accuracy for $S$. This is expected, since we train the $S$-encoder for classification. As for the unspecified component $Z$, our performance on the train and the test set are similar, except for the NORB dataset where our error rate is slightly worse.
For this dataset, the error rate of $S$ in the test set is much larger than that of the train set, and in~\cite{disentanglement} they explain this result by overfitting. Note that for this dataset, there are only five training instances per category, which makes generalization difficult.

\begin{table}[]
\centering
\label{tbl1_numericresults}
\begin{small}
\begin{tabular}{lllllllll}
\toprule
                               & \multicolumn{2}{c}{\textbf{Mnist}} & \multicolumn{2}{c}{\textbf{Sprites}}   \\
\cmidrule{2-3}
\cmidrule{4-5}
                               & Z               & S                & Z                  & S \\ \midrule
Our (train)                          &   87.0\%              &    0.1\%              & 66.0\%               & 0.0\%   \\
Our (test)                           & 87.0\%            & 0.8\%            & 58.0\%               & 0.0\%    \\
~\cite{disentanglement} (train) & -               & -                & 58.6\%               & 5.5\%       \\
~\cite{disentanglement} (test)  & -               & -                & 59.8\%               & 5.2\%       \\
Random-chance                  & \multicolumn{2}{c}{-- 90.0\% --}           & \multicolumn{2}{c}{-- * --}  \\
\bottomrule
~\\
\toprule
                               & \multicolumn{2}{c}{\textbf{NORB}} & \multicolumn{2}{c}{\textbf{Extended-YaleB}} \\
\cmidrule{2-3}
\cmidrule{4-5}
                               & Z               & S                & Z                  & S \\ \midrule
Our (train)                          & 78.9\%          & 1.1\%           & 95.7\%               & 0.00\%                    \\
Our (test)                           & 79.2\%          & 15.2\%          & 96.3\%               & 0.00\%                    \\
~\cite{disentanglement} (train) &  79.8\%          & 2.6\%           & 96.4\%               & 0.05\%               \\
~\cite{disentanglement} (test)  &  79.9\%          & 13.5\%          & 96.4\%               & 0.08\%               \\
Random-chance                  & \multicolumn{2}{c}{-- 80.0\% --}          & \multicolumn{2}{c}{-- 96.4\% --}   \\

\bottomrule
\end{tabular}
\end{small}
\smallskip
\caption{Classification error rate based on $S$ or $Z$ for our model and as reported in~\cite{disentanglement}. *While~\cite{disentanglement} reports 60.7\% chance, we observe 56\% in the test set, and 67\% in the train set.}
\end{table}

\subsection{Financial data}

We applied our method on the daily returns of stocks listed in NASDAQ, NYSE and AMEX, from the Center for Research in Security Prices (CRSP) database. For all datasets, the results were measured on the test set. Specifically, we used daily returns of stocks from the Nasdaq, NYSE and AMEX exchanges. The training set consists of the years 1976-2009 and the test set 2010-2016. Each year is divided into four quarters of approximately $63$ trading days. As an input to the network, we used for each stock the returns of the first $50$ days of each quarter, as well as the  market returns for the same $50$ days. In order to improve generalization, we added $\epsilon_i$ a random noise $N(0,0.0016)$.

The goal of the disentanglement is to separate market behavior from specific stock's movements. In order to do so, we labeled each quarter in the training set differently, so as to have 136 such labels. Next, we let $S$ encode the label information and $Z$ encode  the rest of the information. 



For evaluation, we employed two metrics, (i) checking the stock specific information from $Z$ and (ii) evaluating a trading strategy based on the predictions that came from $Z$. 

For a sanity check, we start by showing that $S$ contains market information. We did a PCA on the $S$-part of the encoding and examined the first component. This component was correlated with the average return of the market during the tested period. The correlation coefficient between the market return on the test period and the first component of the PCA is $0.55$.   



We then defined two stock specific measures based on the Capital Asset Pricing Podel (CAPM)~\cite{sharpe}, which is one of the fundamental quantitative models in finance theory: $\beta$, which is the systematic risk of a given asset, and $\rho$, which is the correlation coefficient with the market during the last year. We constructed a discrete version of these measures with four levels each. The classifier we used is logistic regression, since it dominates the econometrics literature. 
The predictive accuracy on the test set for each of the six models ($2$ measures times $3$ inputs) is given in Table~\ref{Tbl3_betarho}. From this table, we clearly see that we failed to reveal stock properties from $X$ and $S$, but managed to do it from $Z$. 

\begin{table}[]
\centering
\begin{small}
\begin{tabular}{l@{~}l@{~}l@{~}l@{~}l@{~}l}
\toprule
               & \multicolumn{2}{c}{\textbf{beta}} & \multicolumn{1}{c}{\textbf{rho}} & \multicolumn{1}{c}{\textbf{beta}\textbf{~\cite{disentanglement}}} & \multicolumn{1}{c}{\textbf{rho}\textbf{~\cite{disentanglement}}} \\ \midrule
Z              & \multicolumn{2}{c}{35\%}        & \multicolumn{1}{c}{31\%}                           & \multicolumn{1}{c}{31\%}                                 & \multicolumn{1}{c}{30\%}                               \\
S              & \multicolumn{2}{c}{26\%}        & \multicolumn{1}{c}{26\%}                           & \multicolumn{1}{c}{28\%}                                 & \multicolumn{1}{c}{28\%}                               \\
Raw & \multicolumn{2}{c}{26\%}        & \multicolumn{1}{c}{26\%}                           &                                        &                                       \\
Rand  & \multicolumn{2}{c}{25\%}        & \multicolumn{1}{c}{25\%}                           & \multicolumn{1}{c}{25\%}                                 & \multicolumn{1}{c}{25\%}                                \\ \bottomrule
\end{tabular}
\end{small}
\smallskip
\caption{Logistic regression accuracy for $\beta,\rho$}
\label{Tbl3_betarho}
\end{table}

 \begin{table}[t]
\centering
\begin{small}
\begin{tabular}{l@{~}l@{~}l@{~}l@{~}l@{~}l}
\toprule
                              & \multicolumn{1}{c}{\textbf{NY}} & \multicolumn{1}{c}{\textbf{AM}} & \multicolumn{1}{c}{\textbf{NQ}} & \multicolumn{1}{c}{\textbf{All}} & \multicolumn{1}{c}{\textbf{~\cite{disentanglement}}} \\ \midrule
\textbf{Z-1}              & \multicolumn{1}{c}{31\%}                            & \multicolumn{1}{c}{37\%}                            & \multicolumn{1}{c}{30\%}                              & \multicolumn{1}{c}{31\%}                           & \multicolumn{1}{c}{30\%}                                \\
\textbf{S-1}               & \multicolumn{1}{c}{26\%}                            & \multicolumn{1}{c}{24\%}                            & \multicolumn{1}{c}{24\%}                              & \multicolumn{1}{c}{25\%}                           & \multicolumn{1}{c}{27\%}                                \\
\textbf{X-1} & \multicolumn{1}{c}{28\%}                            & \multicolumn{1}{c}{24\%}                            & \multicolumn{1}{c}{24\%}                              & \multicolumn{2}{c}{--- 25\% ---}                                               \\
\textbf{Rnd-1}  & \multicolumn{5}{c}{---------------- 25\% ----------------}                                                                                                                                                             \\ \midrule
\textbf{Z-5}             & \multicolumn{1}{c}{40\%}                            & \multicolumn{1}{c}{49\%}                            & \multicolumn{1}{c}{36\%}                              & \multicolumn{1}{c}{39\%}                           & \multicolumn{1}{c}{34\%}                                \\
\textbf{S-5}             & \multicolumn{1}{c}{26\%}                            & \multicolumn{1}{c}{27\%}                            & \multicolumn{1}{c}{25\%}                              & \multicolumn{1}{c}{26\%}                           & \multicolumn{1}{c}{30\%}                                \\
\textbf{X-5} & \multicolumn{1}{c}{25\%}                            & \multicolumn{1}{c}{29\%}                            & \multicolumn{1}{c}{25\%}                              & \multicolumn{2}{c}{--- 26\% ---}                                               \\
\textbf{Rnd-5}  & \multicolumn{5}{c}{---------------- 25\% ----------------}                                                                                                                                                             \\ \bottomrule
\end{tabular}
\end{small}
 \smallskip
\caption{Logistic regression accuracy for next day/week  volatility. The rightmost column is the results of the model presented in (Mathieu et al. 2016), other columns are the results of our model.}
\label{vol_exc}
\end{table}
\begin{table}
\centering
\begin{small}
\begin{tabular}{l@{~}c@{~}c@{~}c}
\toprule
           & \textbf{Mean} & \textbf{SD} & \textbf{Traded days \%} \\ \midrule
\textbf{Z (Ours)} & 3.1\%          & 0.026       & 89.3\%   \\
\textbf{Z~\cite{disentanglement}} & 2.9\%          & 0.039 & 78.6\%   \\
\textbf{S (Ours)} & 2.4\%          & 0.031       & 82.1\%   \\
\textbf{S~\cite{disentanglement}} & 2.4\%          & 0.028 & 83.2\%   \\
\textbf{X} & 2.6\%          & 0.030        & 78.6\%  \\ \bottomrule                            
\end{tabular}
\end{small}
\smallskip
\caption{Options portfolio returns. The mean, std and percent of trading days with positive returns.}
\label{Tbl_portfoliovol}
\vspace{-5mm}
 \end{table}

A very important measure that is used in options trading is the volatility. Using a model on $Z$, we predicted the next day and next $5$-days volatility. The results are given in Table~\ref{vol_exc}. The accuracy of the different models changes between the stock groups, but the performance is significantly better for the model based on $Z$. 


The volatility is an important component in options pricing models, such as Black-Scholes model~\cite{BS}. We developed the following theoretical options trading strategy: 
(1) We estimated the volatility of a stock based on its volatility in the last fifty trading days. 
(2) We run a classification model for the stock based on either $X$ or $Z$. 
(3) For the ten stocks whose predicted volatility minus measured volatility is the highest, we bought a put and a call option. Similarly for the ten stocks whose predicted volatility minus measured volatility is the lowest, we sold one put option and one call option. The strike price of the options is $5\%$ higher than the current price. The time to expire is $60$ days for the high predicted volatility options and $5$ days for the low volatility ones. 
(4) We cleared position on the next day, i.e., sold options in the case where we bought options yesterday and vice-versa.

Note that this strategy is market neutral and relies only on the volatility. We are aware of the fact that we ignored trading cost, liquidity and other technicalities that make this strategy unrealistic. However, we used it as a way to compare the classifier that used $X$ to the one that used $Z$ as an input. The  results are summarized in Table~\ref{Tbl_portfoliovol}. As one can see, using $Z$ is better. 
The results from~\cite{disentanglement} for financial data are presented next to ours in tables~\ref{Tbl3_betarho},~\ref{vol_exc} and~\ref{Tbl_portfoliovol}. It can be seen that our accuracy and portfolio performance based on $Z$ are better and we also achieved better separation from $S$, since it is almost agnostic to specific stock properties.

\section{Conclusions}
This paper presents an adversarial architecture for solving the problem of disentanglement. Given labeled data, our algorithm encodes it as two separate parts, one that contains the label information and the other that is agnostic to it. We tested the network on visual and financial data, and found that it performed well compared to a leading literature method. Our architecture does not assume a distribution on the unspecified factors and the resultant encoding seemed both more interpretable and more suitable as a representation for learning various unspecified qualities.

\section*{Acknowledgements}

This project has received funding from the European Research Council (ERC) under the European Union's Horizon 2020 research and innovation programme (grant ERC CoG 725974).

\clearpage
{\small
\bibliographystyle{ieee}
\bibliography{gans}

\begin{thebibliography}{10}\itemsep=-1pt

\bibitem{BS}
F.~Black and M.~Scholes.
\newblock The pricing of options and corporate liabilities.
\newblock {\em Journal of Political Economy}, 81(3):637--54, 1973.

\bibitem{infogan}
D.~Y. H. R. S. J. S.~I. Chen, Xi and P.~Abbeel.
\newblock Infogan: Interpretable representation learning by information
  maximizing generative adversarial nets.
\newblock In {\em Advances in neural information processing systems}, pages
  2172--2180, 2016.

\bibitem{denton2015deep}
E.~L. Denton, S.~Chintala, R.~Fergus, et~al.
\newblock Deep generative image models using a laplacian pyramid of adversarial
  networks.
\newblock In {\em Advances in neural information processing systems}, pages
  1486--1494, 2015.

\bibitem{elgammal}
A.~Elgammal and C.-S. Lee.
\newblock Separating style and content on a nonlinear manifold.
\newblock In {\em Computer Vision and Pattern Recognition, 2004. CVPR 2004.
  Proceedings of the 2004 IEEE Computer Society Conference on}, volume~1, pages
  I--I. IEEE, 2004.

\bibitem{fletch}
L.~A. Fletcher and R.~Kasturi.
\newblock A robust algorithm for text string separation from mixed
  text/graphics images.
\newblock {\em IEEE transactions on pattern analysis and machine intelligence},
  10(6):910--918, 1988.

\bibitem{yaleB}
A.~S. Georghiades, P.~N. Belhumeur, and D.~J. Kriegman.
\newblock From few to many: Illumination cone models for face recognition under
  variable lighting and pose.
\newblock {\em IEEE transactions on pattern analysis and machine intelligence},
  23(6):643--660, 2001.

\bibitem{gan}
I.~Goodfellow, J.~Pouget-Abadie, M.~Mirza, B.~Xu, D.~Warde-Farley, S.~Ozair,
  A.~Courville, and Y.~Bengio.
\newblock Generative adversarial nets.
\newblock In {\em NIPS}, pages 2672--2680. 2014.

\bibitem{huang}
F.~J. Huang, Y.-L. Boureau, Y.~LeCun, et~al.
\newblock Unsupervised learning of invariant feature hierarchies with
  applications to object recognition.
\newblock In {\em Computer Vision and Pattern Recognition, 2007. CVPR'07. IEEE
  Conference on}, pages 1--8. IEEE, 2007.

\bibitem{adam}
D.~Kingma and J.~Ba.
\newblock Adam: A method for stochastic optimization.
\newblock In {\em The International Conference on Learning Representations
  (ICLR)}, 2016.

\bibitem{kingma}
D.~P. Kingma, S.~Mohamed, D.~J. Rezende, and M.~Welling.
\newblock Semi-supervised learning with deep generative models.
\newblock In {\em Advances in Neural Information Processing Systems}, pages
  3581--3589, 2014.

\bibitem{fader}
G.~Lample, N.~Zeghidour, N.~Usunier, A.~Bordes, L.~Denoyer, and M.~Ranzato.
\newblock Fader networks: Manipulating images by sliding attributes.
\newblock {\em CoRR}, abs/1706.00409, 2017.

\bibitem{mnist}
Y.~LeCun and C.~Cortes.
\newblock {MNIST} handwritten digit database.
\newblock 2010.

\bibitem{norb}
Y.~LeCun, F.~J. Huang, and L.~Bottou.
\newblock Learning methods for generic object recognition with invariance to
  pose and lighting.
\newblock In {\em Computer Vision and Pattern Recognition, 2004. CVPR 2004.
  Proceedings of the 2004 IEEE Computer Society Conference on}, volume~2, pages
  II--104. IEEE, 2004.

\bibitem{disentanglement}
M.~F. Mathieu, J.~J. Zhao, J.~Zhao, A.~Ramesh, P.~Sprechmann, and Y.~LeCun.
\newblock Disentangling factors of variation in deep representation using
  adversarial training.
\newblock In D.~D. Lee, M.~Sugiyama, U.~V. Luxburg, I.~Guyon, and R.~Garnett,
  editors, {\em Advances in Neural Information Processing Systems 29}, pages
  5040--5048. Curran Associates, Inc., 2016.

\bibitem{radford}
A.~Radford, L.~Metz, and S.~Chintala.
\newblock Unsupervised representation learning with deep convolutional
  generative adversarial networks.
\newblock {\em arXiv preprint arXiv:1511.06434}, 2015.

\bibitem{sprites}
S.~E. Reed, Y.~Zhang, Y.~Zhang, and H.~Lee.
\newblock Deep visual analogy-making.
\newblock In {\em Advances in Neural Information Processing Systems}, pages
  1252--1260, 2015.

\bibitem{sharpe}
O.~W. Sharpe and M.~Miller.
\newblock Capm.
\newblock {\em Equilibrium}, 1964.

\bibitem{tenenbaum}
J.~B. Tenenbaum and W.~T. Freeman.
\newblock Separating style and content.
\newblock {\em Advances in neural information processing systems}, pages
  662--668, 1997.

\end{thebibliography}
}

\end{document}